\documentclass[10pt,twocolumn,letterpaper]{article}

\usepackage{cvpr}
\usepackage{times}
\usepackage{epsfig}
\usepackage{graphicx}
\usepackage{amsmath}
\usepackage{amssymb}
\usepackage{setspace}
\usepackage{siunitx}
\usepackage{float}
\usepackage{blindtext}
\usepackage{graphicx}
\usepackage{caption}

\usepackage{authblk}
\usepackage{abstract}

\usepackage[pagebackref=true,breaklinks=true,letterpaper=true,colorlinks,bookmarks=false]{hyperref}

\DeclareMathOperator*{\argmin}{\arg\!\min}

\cvprfinalcopy 

\def\cvprPaperID{2633} 

\ifcvprfinal\fi
\begin{document}

\title{Sparse Photometric 3D Face Reconstruction Guided by Morphable Models}

\newcommand\CoAuthorMark{\footnotemark[\arabic{footnote}]}
\newcommand*\samethanks[1][\value{footnote}]{\footnotemark[#1]}
\renewcommand\Authands{ and }

\author[1]{Xuan Cao\thanks{These authors contribute to the work equally.}}
\author[1]{Zhang Chen\samethanks}
\author[1]{Anpei Chen}
\author[1]{Xin Chen}
\author[1]{Cen Wang}
\author[1]{Jingyi Yu}
\affil[1]{ShanghaiTech University, Shanghai, China. \texttt{\{caoxuan, chenzhang, chenap, chenxin2, wangcen, yujingyi\}@shanghaitech.edu.cn}}


\twocolumn[{%
\renewcommand\twocolumn[1][]{#1}%
\maketitle
\begin{center}
\centering
\includegraphics[width=1\textwidth]{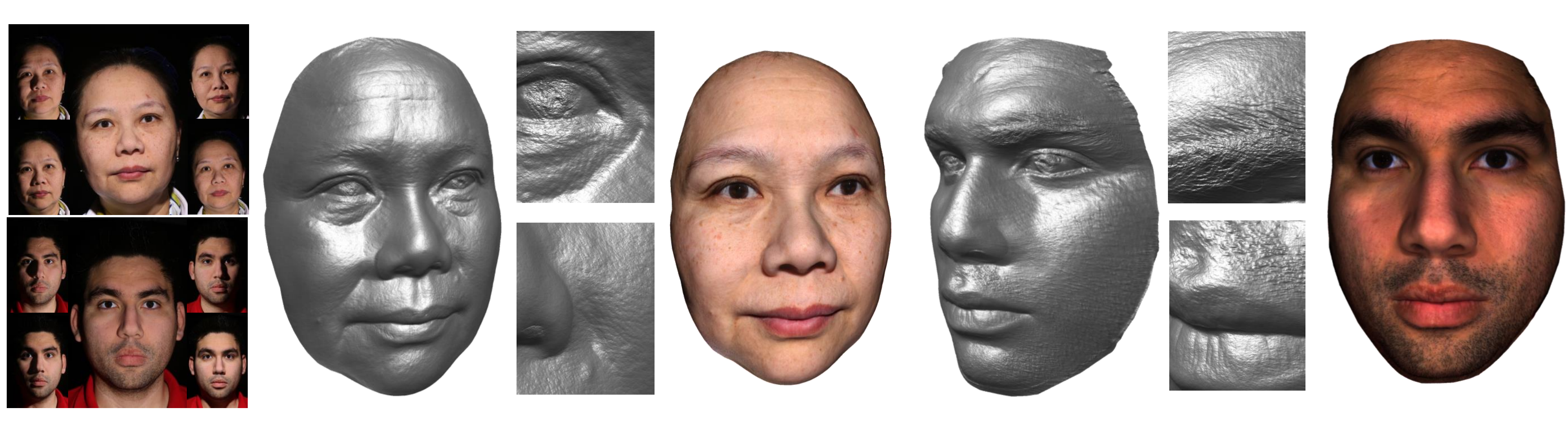}
\captionof{figure}{Sample results using our sparse PS reconstruction. By using just 5 input images (left), our method can recover very high quality 3D face geometry with fine geometric details.}\label{fig:Teaser}
\end{center}%
}]

\saythanks




\begin{abstract}
We present a novel 3D face reconstruction technique that leverages sparse photometric stereo (PS) and latest advances on face registration/modeling from a single image. We observe that 3D morphable faces approach\cite{huber2016multiresolution} provides a reasonable geometry proxy for light position calibration. Specifically, we develop a robust optimization technique that can calibrate per-pixel lighting direction and illumination at a very high precision without assuming uniform surface albedos. Next, we apply semantic segmentation on input images and the geometry proxy to refine hairy vs. bare skin regions using tailored filters. Experiments on synthetic and real data show that by using a very small set of images, our technique is able to reconstruct fine geometric details such as wrinkles, eyebrows, whelks, pores, etc, comparable to and sometimes surpassing movie quality productions.
\end{abstract}

\section{Introduction}

The digitization of photorealistic\thanks{432423} 3D face is a long-standing problem and can benefit numerous applications, ranging from movie special effects\cite{Alexander2009Creating} to face detection and recognition\cite{hassner2016pooling}. Human faces contain both low-frequency geometry (e.g., nose, cheek, lip, forehead) and high-frequency details (e.g., wrinkles, eyebrows, beards, and pores). Passive reconstruction techniques such as stereo matching\cite{hirschmuller2008stereo}, multiview geometry\cite{Joo_2015_ICCV}, structure-from-motion\cite{bartoli2005structure}, and most recently light field imaging\cite{raytrixHomePage} can now reliably recover low frequency geometry. Recovering high-frequency details is way more challenging. Successful solutions still rely on professional capture systems such as 3D laser scans or ultra-high precision photometric stereo such as the USC Light Stage systems\cite{Ghosh2011Multiview,Ma2007Rapid}. Developing commodity solutions to simultaneously capture low-frequency and high-frequency face geometry is particularly important and urgent.

To quickly reiterate the challenges, PS requires knowing the lighting direction at a very high precision. It is common practice to position a point light at a far distance to emulate a directional light source for easy calibration. In reality, such setups are huge and require strong lighting power. Alternatively, one can use near-field point light sources \cite{Smith2005The,B2010Effects} to set up a more portable system. However, calibrating the lighting direction becomes particularly difficult: one needs to know both the position of the light source(s) and the face geometry. The former can be estimated by using sphere \cite{takai2009difference,wong2008recovering,schnieders2013camera} or planar light probes \cite{park2014calibrating,visentini2015simultaneous}. The latter, however, is precisely what the initial problem aims to resolve and therefore the problem is ill-posed and relies on additional priors.

We leverage recent advances on morphable 3D faces for pose estimation and geometric reconstruction techniques  \cite{Booth2016A,huber2016multiresolution,Roth2016Adaptive,Sinha2017SurfNet,Cole2017Synthesizing,Dou2017End-To-End,Trigeorgis2017Face,Jackson2017Large,Richardson2017Learning}. Such solutions only use very few or even a single image as input, frontal parallel or side views.  The 3D face model can then be inferred by morphing the canonical model. Their results are impressive for neutral facial expressions \cite{Booth2016A,Cole2017Synthesizing}. But the high frequency details are still largely missing \cite{Bolkart2016A,Roth2016Adaptive,Piotraschke2016Automated}. The seminal work of \cite{Jackson2017Large,Richardson2017Learning} manages to recover high frequency geometry to some extent but the results are still not comparable to high-end solutions (e.g., from the USC Light Stage \cite{Ghosh2011Multiview,Ma2007Rapid}).

In this paper, we combine morphable face approach with sparse PS for ultra high quality 3D face reconstruction. We observe that the morphable face approach\cite{huber2016multiresolution} provides a reasonable geometry proxy for light position calibration. Specifically, we develop a robust optimization technique that can calibrate per-pixel incident lighting direction as well as lighting illumination at a very high precision. Our technique overcomes the artifacts of geometric deformations caused by inaccurate lighting estimation and produces a high-precision normal map. Next, we apply semantic segmentation on input images and the approximated geometry to separately refine hairy vs. bare skin regions. For hairy regions, we adopt a bidirectional extremum filter for detail-preservation smoothing. Comprehensive experiments on synthetic and publicly available datasets demonstrate our approach is reliable and accurate. For real data, we construct a capture dome composed of 5 near point light sources with an entry-level DSLR camera. Our technique is able to deliver high quality reconstructions with ultra-fine geometric details such as wrinkles, eyebrows, whelks, pores etc. The reconstruction quality is comparable to and sometimes surpasses movie quality productions based on dense inputs and expensive setups.

\section{Related Works}
3D face reconstruction has a long history in computer vision. The literature is huge and we only discuss the most relevant ones to our approach.

\paragraph{Photometric Stereo.} In computer graphics and vision, photometric Stereo (PS) \cite{Woodham1980Photometric} is the widely adopted technique for inferring the normal map of the face. The normal map can then be integrated (e.g, using Poisson completion\cite{simchony1990direct}) to reconstruct the point cloud and then mesh. We refer the readers to the comprehensive survey \cite{Herbort2011An} for the benefits and problems of the state-of-the-art methods. In general, recovering high quality 3D geometry requires using complex setups. The most notable work is the USC Light Stage\cite{Ma2007Rapid,Ghosh2011Multiview} that utilizes 156 dedicatedly controlled light sources simulating the first-order spherical harmonics function. Their solution can produce very high-quality normal map using near point light sources and the results are superb and have been adopted in movie productions. The setup, however, is rather expensive in cost and labor. Developing cheaper solutions capable of producing similar quality reconstruction is highly desirable , but by far few solutions can match the Light Stage.

\paragraph{2D-to-3D Conversion.} There is an emerging interest on directly converting a 2D face image to a 3D face model. Most prior works can be categorized into 3D-morphable faces and learning-based techniques. Booth et al. \cite{Booth2016A} automatically synthesized a 3D morphable model from over 10,000 3D faces. Bolkar \cite{Bolkart2016A} utilized a multiline model based learning framework that uses much smaller training datasets. \cite{huber2016multiresolution} proposed a Surrey Face Model which provides high resolution 3D morphable model and landmarks alignment. Face models obtained from these approaches are sensitive to pose, expression, illumination, etc, and the problem can be mitigated by using more images \cite{Roth2016Adaptive,Piotraschke2016Automated} or special facial feature decoders \cite{Cole2017Synthesizing}.

In the past few years, a large volume of deep learning based approaches have shown great success on face pose and geometry estimations \cite{Dou2017End-To-End,Jackson2017Large,Richardson2017Learning}. Trigeorgis et al. \cite{Trigeorgis2017Face} tailored a deep CNN to estimate face normal map 'in the wild' and then inferred the face shape. Tran et al. \cite{Tran2017Regressing} applied regression to recover discriminative 3D morphable face models. The main goal of these approaches is face recognition and the recovered geometry is generally highly smooth. Most recent techniques\cite{Richardson2017Learning} can recover certain medium-scale details such as deep wrinkles but the quality is still not comparable to professional solutions.

In a similar vein as ours, Lu et al. \cite{Lu2013A} combined a low resolution depth with high resolution photometric stereo where the depth map is obtained via structured light. Compared with the structured light results that are highly noisy, the morphable 3D face geometry is smoother but less accurate. We further conduct optimization and semantic segmentations for refinement.

\begin{figure*}[t]
\centering
\includegraphics[width=1\textwidth]{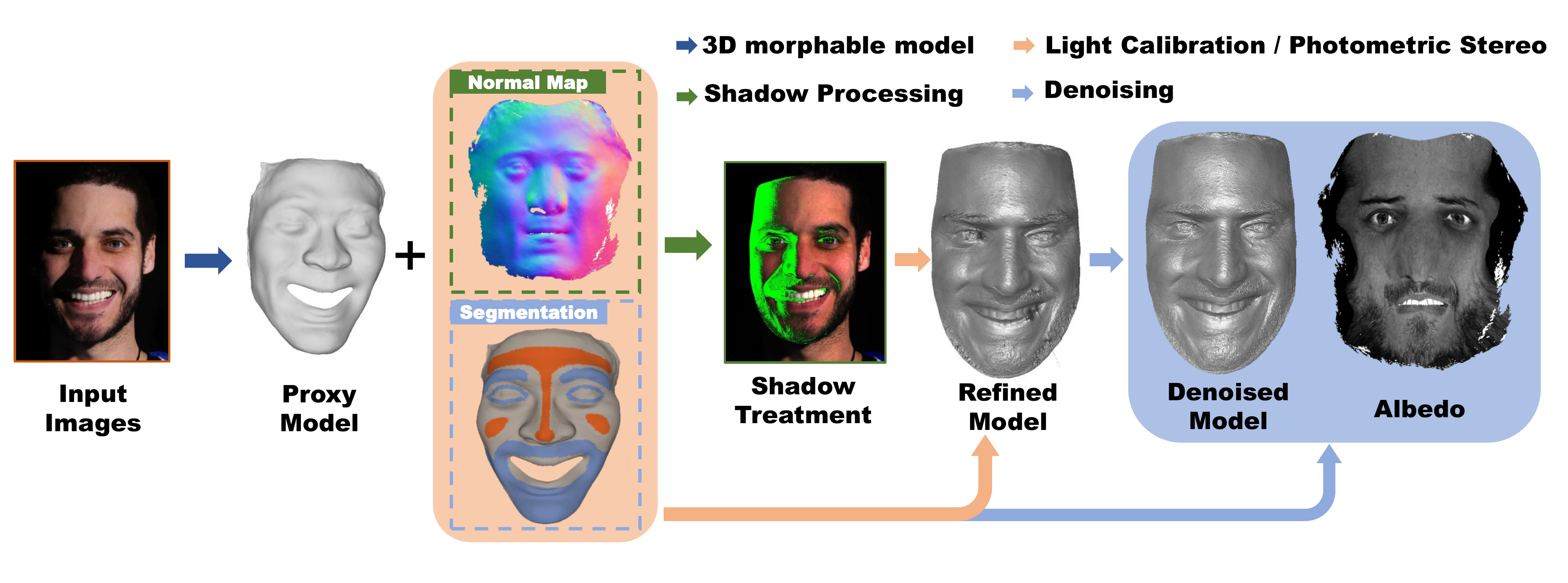}
\caption{The processing pipeline of our proposed sparse PS face reconstruction framework.}\label{fig:FullPipeline}
\end{figure*}

\section{Lighting Calibration}

We aim to replace distant directional light sources with near point light sources, to substantially reduce the cost and space requirement of the PS setup while maintaining the performance. The key challenge is the requirement of estimating relative positions from each point light to surface points. In addition, illumination variations across the light sources can cause severe geometry deformation\cite{B2010Effects}. In this section, we describe a robust auto-calibration technique that conducts estimation for the positions and illumination of near point lights.

Fig. \ref{fig:FullPipeline} shows our processing pipeline. We first obtain a proxy face model through the 3D morphable model, with pose and expression aligned with the input images. We then retrieve the normal and positions at surface vertices points and develop an optimization scheme that, without assuming uniform albedo, jointly estimates positions and illumination of all lights.


\subsection{Shading Model and Proxy Geometry}\label{section:Proxy}
\begin{figure}[t]
\centering
\includegraphics[width=0.5\textwidth]{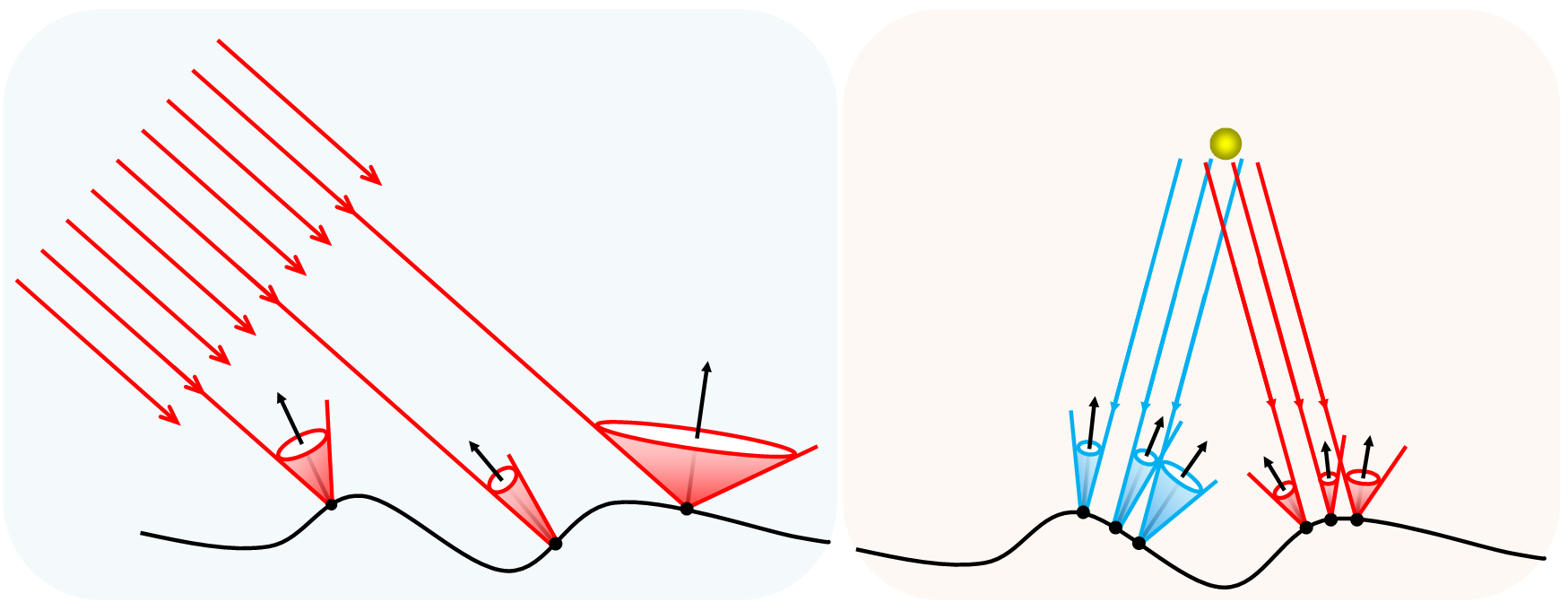}
\caption{In traditional PS, parallel (left) and point light (right) calibrations rely on the uniform albedo assumption.}\label{fig:TwoCones}
\end{figure}

Under the Lambertian assumption, the intensity of a pixel is:
\begin{equation}\label{lambertian_vector}
I = \rho N\cdot L,
\end{equation}
where $\rho$ and $N$ are the albedo and normal at the pixel, $L$ is the light direction at the corresponding 3D point. Each point maps to a triplet of ($I$, $\rho$, $N$), and defines a cone of potential lighting directions as in Fig. \ref{fig:TwoCones}. For the directional light source model, three linearly independent triplets of ($I$, $\rho$, $N$) can be used to estimate the light direction. For near point light model, however, it is critical to know both the positions of the light source and the vertex.

We exploit the recent 3DMM \cite{huber2016multiresolution} to generate a proxy face model at first. However, it is important to note that the model obtained from these techniques are not directly applicable for high quality 3D reconstruction: 1) the resulting face model does not match the captured image, especially near the silhouettes; 2) the predicted model generally exhibits neutral expressions but we aim to capture a much richer class of expressions; 3) the face model is incomplete: it lacks facial regions such as the upper forehead and mouth cavity.

In our approach, we use the proxy face model to first estimate the surface normal and calibrate the light sources. We also use the proxy geometry to segment the photographed face into two categories of regions: smooth regions including lower forehead, cheekbone, inner cheek and nose bridge are potentially suitable for reliable normal estimations, and hairy regions including eyebrows, eyelids, mouth surroundings, chin and outer cheek, are generally noisy and require additional processing. Since the proxy face models from different images share vertices and connectivities, we can conduct coherent segmentations in all input images.


\subsection{Near Point Light Calibration}
We employ the proxy model to calibrate light positions while accounting for illumination variation. There are two classical approaches for calibrating near point lights. One resorts to spherical probes\cite{wong2008recovering,schnieders2013camera} or planar light probes \cite{park2014calibrating,visentini2015simultaneous}. These light probe-based methods recover the light positions in camera coordinate system. The second utilizes the reflectance data of the Lambertian surface with known geometry. For example, \cite{zheng1991estimation} recovers the light directions from known surface normals at multiple points with a uniform albedo. By further assuming that neighboring pixels have similar albedo, \cite{mancini19923} estimates the light directions at multiple vertexes and subsequently the light positions. \cite{weber2001practical} uses two cubes covered with white paper for light position calibration. All these approaches require either extra instruments or uniform albedo assumptions.

\begin{figure}[h]
\centering
\includegraphics[width=0.4\textwidth, height=0.2\textheight]{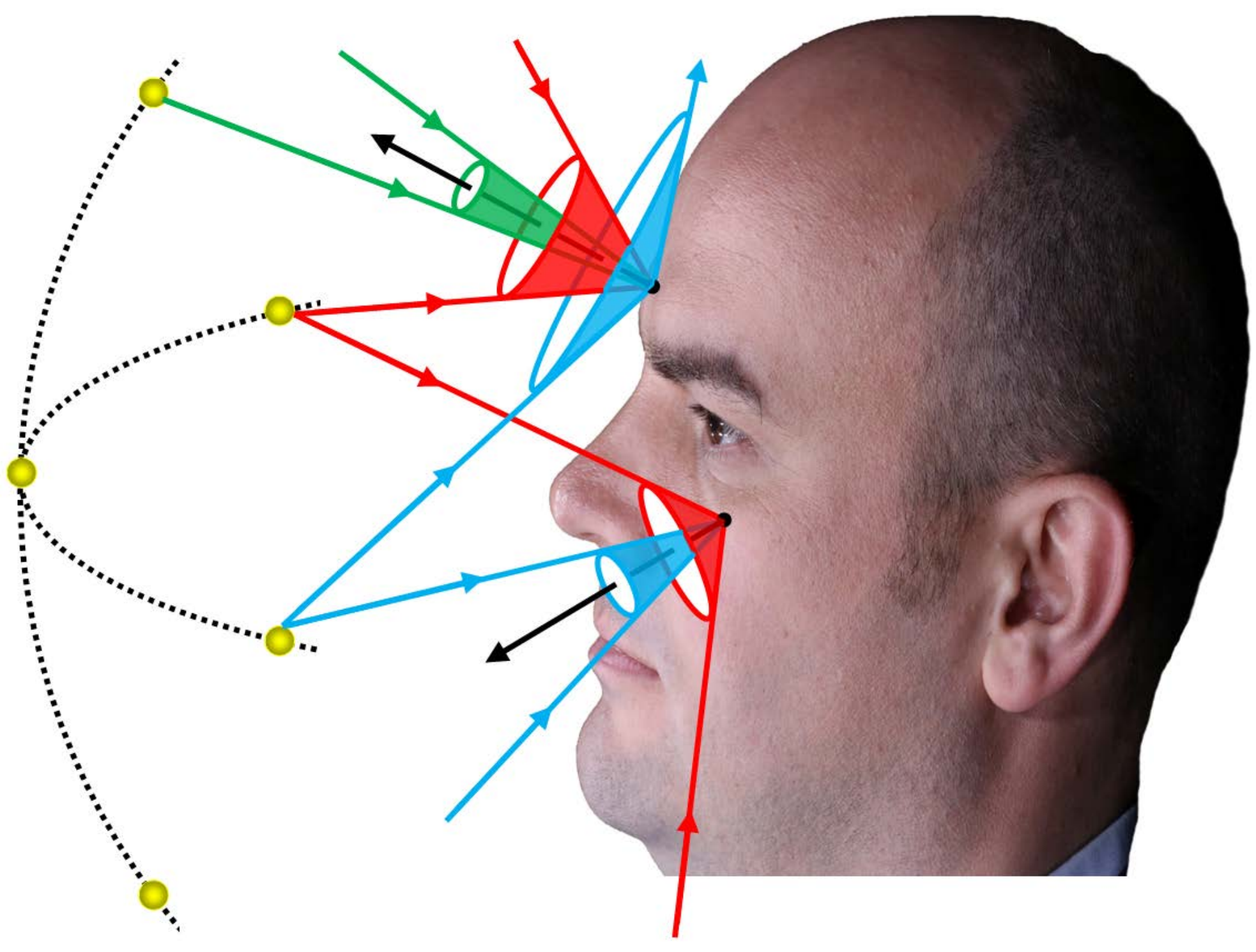}
\caption{Our light calibration approach uses the proxy model for light position calibrations. By assuming known surface normal, we can form over-determined linear systems using multiple light sources on key surface. Cones of different colors correspond to constraints imposed by different light sources.}\label{fig:JointCalibration}
\end{figure}

Our approach is instrument-free and does not make uniform albedo assumption.  We first extract $m$ key points from the smooth regions in the semantically segmented face images and obtain their normal $\mathbf{N}\in\mathbb{R}^{m\times 3}$ along with their corresponding vertex positions $\mathbf{V}\in\mathbb{R}^{m\times 3}$.  Recall, for non-uniform albedos, we will not be able to solve for $\rho$ and $L$ in Equation \eqref{lambertian_vector} separately for each light. We therefore jointly solve Equation \eqref{lambertian_vector} for all $m$ key points and $n$ lights as shown in Fig. \ref{fig:JointCalibration}.

Recall under the directional lighting model, we have
\begin{equation}\label{lambertian_matrix}
\textbf{I}=\text{diag}(\boldsymbol{\rho})\mathbf{N}\mathbf{L}^\mathsf{T},
\end{equation}
where $\mathbf{I}\in\mathbb{R}^{m\times n}$ is the image intensity at the key points and  $\mathbf{L}\in\mathbb{R}^{n\times 3}$ is the lighting directions. For near point lights with inconsistent illumination, we replace $\mathbf{L}$ with the scaled directions $\mathbf{D}_{i, j}$ for the $j^{th}$ light and the $i^{th}$ key point so that:
\begin{equation}\label{Dij}
\begin{split}
&\mathbf{D}_{i, j} = \boldsymbol{\beta}_{j}\cdot\frac{1}{\left\lVert\mathbf{P}_{j}-\mathbf{V}_{i}\right\rVert_2^2}\cdot\frac{(\mathbf{P}_{j}-\mathbf{V}_{i})}{\left\lVert\mathbf{P}_{j}-\mathbf{V}_{i}\right\rVert_2}=\frac{\boldsymbol{\beta}_j(\mathbf{P}_{j}-\mathbf{V}_{i})}{\left\lVert\mathbf{P}_{j}-\mathbf{V}_{i}\right\rVert_2^3}, \\
&\text{ for }i = 1, 2,...,m\text{ and }j=1, 2,...,n.
\end{split}
\end{equation}
where $\boldsymbol{\beta}\in\mathbb{R}^{n\times 1}$, $\mathbf{P}\in\mathbb{R}^{n\times3}$ are the illumination and positions of all lights. The second term $\frac{1}{\left\lVert\mathbf{P}_{j}-\mathbf{V}_{i}\right\rVert_2^2}$ reflects the inverse square law between light brightness and distance which  is critical for near point light calibrations. The image intensity of $i^{th}$ key point under the $j^{th}$ lighting is represented as
\begin{equation}\label{lambertian_perpixel}
\mathbf{I}_{i, j}=\boldsymbol{\rho}_{i}\mathbf{N}_{i}\mathbf{D}_{i, j}^\mathsf{T}.
\end{equation}



To solve for the illumination $\boldsymbol{\beta}$ and position $\mathbf{P}$ of light sources, we formulate the estimation as the following optimization problem:
\begin{equation}\label{optim_initial}
\begin{split}
\tilde{\boldsymbol{\rho}}, \tilde{\boldsymbol{\beta}}, \tilde{\mathbf{P}}=\argmin_{\boldsymbol{\rho}, \boldsymbol{\beta}, \mathbf{P}} &\sum_{i=1}^{m}\sum_{j=1}^{n}\left\lVert\mathbf{I}_{i, j}-\boldsymbol{\rho}_{i}\mathbf{N}_{i}\mathbf{D}_{i, j}^\mathsf{T}\right\rVert_2^2\\
&+\lambda_{1}\sum_{j=1}^n\left\lVert\bar{\boldsymbol{\beta}}-\boldsymbol{\beta}_j\right\rVert_2^2+\lambda_{2}\left\lVert\boldsymbol{\rho}\right\rVert_2^2\\
&+\lambda_3\sum_{j=1}^{n}(\left\lVert\mathbf{P}_j\right\rVert_2-d)_2^2,
\end{split}
\end{equation}
where $\bar{\boldsymbol{\beta}}$ is the mean of all elements in $\boldsymbol{\beta}$, $d\in\mathbb{R}$ is a prior of the distance between the lights and geometry proxy. The first term represents the least square error under the Lambertian surface model. The second term is based on the fact that illumination variations are relatively small under our setup.

Note that there is a scale ambiguity between $\boldsymbol{\rho}$ and $\boldsymbol{\beta}$ in Equation \eqref{optim_initial}. Therefore we append the third term to enforce the uniqueness of $\boldsymbol{\rho}$ and $\boldsymbol{\beta}$.  The last term aims to remove outliers in $\mathbf{I}$, e.g., the ones deviate greatly from the Lambertian surface model due to noise.
We initialize $\boldsymbol{\rho}$ as the maximal image intensity of each key point across the light sources. We initialize $\boldsymbol{\beta}$ as vector $\mathbf{1}$, and $\mathbf{P}$ on the half sphere with radius $d$ centered at the origin with a positive $z$.

Since the normal from the proxy face model may be inaccurate, the $\tilde{\boldsymbol{\beta}}$ and $\tilde{\mathbf{P}}$ estimated from Equation \eqref{optim_initial} are impeded by the error in $\mathbf{N}$. To compensate for this, we further refine the estimates by iterating between the following two optimizations:
\begin{itemize}
\item Fix the estimated illumination $\boldsymbol{\beta}$ and positions $\mathbf{P}$ of all lights, update albedo $\boldsymbol{\rho}$ and normal $\hat{\mathbf{N}}$ of the key points in Equation \eqref{optim_step1},
\begin{equation}\label{optim_step1}
\begin{split}
\min_{\boldsymbol{\rho}, \hat{\mathbf{N}}} &\sum_{i=1}^{m}\sum_{j=1}^{n}\left\lVert\mathbf{I}_{i, j}-\boldsymbol{\rho}_{i}\hat{\mathbf{N}}_{i}\mathbf{D}_{i, j}^\mathsf{T}\right\rVert_2^2
+\lambda_{n}\left\lVert\hat{\mathbf{N}}-\mathbf{N}\right\rVert_2^2.
\end{split}
\end{equation}
\item Fix the estimated albedo $\boldsymbol{\rho}$ and normal $\hat{\mathbf{N}}$ for the keypoints, update illumination $\boldsymbol{\beta}$ and positions $\mathbf{P}$ of all lights in Equation \eqref{optim_step2},
\begin{equation}\label{optim_step2}
\begin{split}
\min_{\boldsymbol{\beta}, \mathbf{P}} &\sum_{i=1}^{m}\sum_{j=1}^{n}\left\lVert\mathbf{I}_{i, j}-\boldsymbol{\rho}_{i}\hat{\mathbf{N}}_{i}\mathbf{D}_{i, j}^\mathsf{T}\right\rVert_2^2\\
&+\lambda_{\beta}\sum_{j=1}^n\left\lVert\bar{\boldsymbol{\beta}}-\boldsymbol{\beta}_j\right\rVert_2^2
+\lambda_{P}\sum_{j=1}^{n}(\left\lVert\mathbf{P}_j\right\rVert_2-d)_2^2.
\end{split}
\end{equation}
\end{itemize}
For our experiments, we empirically set $\lambda_1=\lambda_2=\lambda_{\beta}=0.001$, $\lambda_3=\lambda_{P}=0.0001$ and $\lambda_n=10^{-6}$.

\subsection{Handling Shadow Areas}
Human faces contain non-convex geometry in multiple regions, such as the surroundings of nose and eye sockets. Image intensity in these shadow areas clearly violates the Lambertian surface model and significantly degrades normal estimations, especially with insufficient number of lighting directions. Solutions to detect shadow areas, such as intensity-based segmentation, are sensitive to the image content, especially with non-uniform albedos.

We observe that the proxy face model provides crucial cues to eliminate the impact of shadows. Denote $\Lambda$ as the set of pixels inside the reconstructed regions. For pixel $i\in\Lambda$ and light $j\in\{1, 2, ..., n\}$, from Equation \eqref{lambertian_perpixel}, we have
\begin{equation}\label{shadow_rho}
\boldsymbol{\rho}_{i} = \frac{\mathbf{I}_{i, j}}{\mathbf{N}_{i}\mathbf{D}_{i, j}^\mathsf{T}}.
\end{equation}
We already have $\mathbf{N}_i$ from the proxy face model and we can compute $\mathbf{D}_{i, j}^\mathsf{T}$  by substituting estimated $\boldsymbol{\beta}$ and $\mathbf{P}$ into Equation \eqref{Dij}. Therefore, we can calculate $\boldsymbol{\rho}_{i, j}$ as the albedo of pixel $i$ under light $j$. Conceptually, pixel $i$ is in shadow under light $j$ if $\boldsymbol{\rho}_{i, j}=0$. In reality, $\boldsymbol{\rho}_{i, j}$, however, may be nonzero even when pixel $i$ lies in shadow due to calibration errors, inter-reflections, subsurface scattering, etc.

We develop a simple but effective technique for handling shadows. Equation (\ref{equ:413}) reveals that, for each pixel $i\in\Lambda$, we can first calculate the mean albedo $\bar{\boldsymbol{\rho}_{i}}$ of this pixel and obtain the set of $\boldsymbol{\rho}_{i, j}$ higher than $\bar{\boldsymbol{\rho}_{i}}$. We then calculate the mean value $\mu_i$ of the selected set and deem a pixel $i$ out of shadow under light $j$ if the calculated albedo is higher than $(1-\tau)\mu_i$, where $\tau$ is set as $0.4$ in our experiments. For the normal estimation at pixel $i$, we then only use the remaining lights in $\mathcal{L}_i$.
\begin{equation}\label{equ:413}
\left\{
\begin{split}
& \mathcal{S}_i=\{\boldsymbol{\rho}_{i, j}\mid\boldsymbol{\rho}_{i, j}>\bar{\boldsymbol{\rho}_{i}}\} \\
& \mu_i=mean(\mathcal{S}_i) \\
& \mathcal{L}_i=\{j\mid\boldsymbol{\rho}_{i, j}>(1-\tau)\mu_i\}
\end{split}
\right.
\end{equation}
We further deem lights whose incident lighting direction is larger than $\ang{90}$ invalid, as shown in Eq.(\ref{equ:414}). Consequently, for pixel $i$, we only use valid light sources $\mathcal{V}_i$ to estimate the normal at this pixel.
\begin{equation}\label{equ:414}
\left\{
\begin{split}
& \mathcal{A}_i=\{j\mid\mathbf{N}_{i}\mathbf{D}_{i, j}^\mathsf{T}>0\} \\
& \mathcal{V}_i= \mathcal{L}_i \cap \mathcal{A}_i
\end{split}
\right.
\end{equation}

\subsection{Denoising Hairy Regions}
Hairy regions of the face such as shaggy beards and bushy eyebrows contain very complex geometry and shading effects where the Lambertian model fails. Under sparse lightings, normal estimations in these regions are particularly noisy.

Given $N_x$, $N_y$, $N_z$ as the $x$, $y$, $z$ components of the normal, we first compute depth gradient maps $G_x$, $G_y$ as
\begin{equation}\label{pq}
G_x=-\frac{N_x}{N_z},\quad G_y=-\frac{N_y}{N_z}.
\end{equation}
Our goal is to denoise the gradient map. However, traditional denoising filters also remove high-frequency geometry. We adopt a simple yet effective bidirectional extremum filter \ref{equ:BidirectionalExtremumFilter} to eliminate the singular values in gradient maps while preserving high-frequency geometry. Specifically, we first center the gradient map $G$ (either $G_x$ or $G_y$) to have a zero mean and take its element-wise absolute value to compute a transformed gradient map $G^t$. If the transformed gradient $G^{t}_{uv}$ exceeds the mean of transformed gradient map $\bar{G^t}$ scaled by a factor $\sigma$, we replace the original gradient $G_{uv}$ with the median of the neighboring gradients in $G$.

\begin{small}
\begin{equation}\label{equ:BidirectionalExtremumFilter}
\left\{
\begin{split}
& G^{t}=\left|G-\bar{G}\right|\\
& G_{uv}^{'}=
\begin{cases}
median(G_{{(k,l)}\in win(u, v)}),&G^{t}_{uv}>\sigma\bar{G^{t}}\\
G_{uv},&G^{t}_{uv}\leq \sigma\bar{G^{t}}
\end{cases}\\
\end{split}
\right.
\end{equation}
\end{small}
\noindent
where $win(u, v)$ is the neighboring area around pixel at $(u, v)$. In our experiments, In our experiments, we set $\sigma$ as $5$ and neighboring area as $10\times10$, and apply the filter to the hairy regions.

\paragraph{Iterative Optimization.} Once we obtain the face model after all the steps mentioned above, we can substitute the obtained high quality model into the lighting calibration modules and repeat the process for further refinement, as shown in Fig. \ref{fig:FullPipeline}. The process stops when the change of estimation is rather small. In all experiments in the paper, we iteratively conduct the process no more than 10 times and the results are already highly accurate.


\section{Experiments}
We have conducted comprehensive experiments on both publicly available datasets and our own captured data.

\subsection{Synthetic Data}
For synthetic experiments, we use the face models reconstructed from the Light Stage \cite{Ma2007Rapid} as the ground truth. The model contains highly accurate low-frequency geometry and high-frequency details. Using the Lambertian surface model and point light source model, we render $5$ images of the model illuminated by different point light sources on a sphere surround the face. The radius of the sphere is set to be equal to the distance between the forehead and chin. We use the rendered data to compare the accuracy of various reconstruction schemes.

\begin{figure}[t]
\centering
\includegraphics[width=0.45\textwidth]{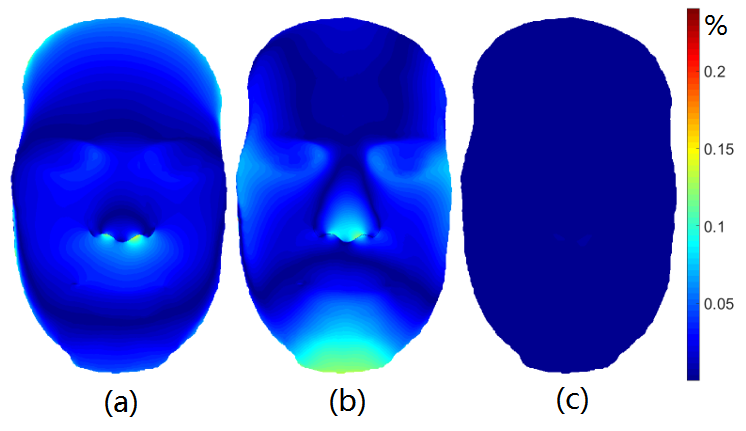}
\caption{Reconstruction error comparisons. (a) Parallel lighting assumption with known albedo and normal. (b) Using matrix-factorization \cite{park2017robust}. (c) Our approach. Error is measured in terms of the ratio between the depth deviation to the ground truth depth.}\label{fig:DepthError}
\end{figure}

We first test the parallel light assumption. Specifically, we analyze two scenarios: 1) using the ground truth albedo and normal to compute parallel light directions, and then using the light directions to calculate the normal, and 2) using the matrix-factorization-based method \cite{park2017robust} to simultaneously solve for parallel light directions and normal. For point light model, we use a proxy face model predicted from one of the rendered image for lighting calibration and use the results to obtain per-pixel light directions.

To apply \cite{park2017robust}, we use $5$ input images with the normal from proxy model as prior. To measure the reconstruction error, we align the reconstructed face models with ground truth model under the same scale and then calculate the reconstruction error as the per-pixel sum of the absolute depth error normalized by the depth range of ground truth model. Fig. \ref{fig:DepthError} shows the face models reconstructed using parallel light model yield noticeable geometric deformations while the face model from our method produces much smaller error. Notice that all three face models uniformly incur larger errors around the forehead and lower edge of nose tip. This is because at such spots, $N_z$ approaches $0$, and according to Equation \eqref{pq}, a small disturbance in normal incurs large errors in $G_x$ and $G_y$ and subsequently the depth estimation.

\begin{figure}[t]
\centering
\includegraphics[width=0.45\textwidth]{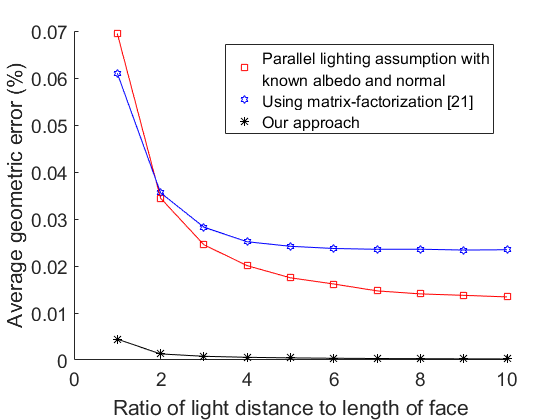}
\caption{Reconstruction errors under different light distances using our technique vs. the state-of-the-art. Unit distance corresponds to the face length (the distance from forehead and chin). }\label{fig:DepthErrorAlongDistance}
\end{figure}

We further test how parallel lighting approximations vary when the lights are positioned farther away. We vary the distance between light sources and the face, ranging from one unit of the distance between the forehead and chin to ten units, as shown in Fig. \ref{fig:DepthErrorAlongDistance}. The error decreases as the distance is farther away, for both parallel and point light source models. However, our method uniformly outperforms the other two with a significant margin.



\subsection{Real Data}
\begin{figure}[h]
\centering
\includegraphics[width=0.45\textwidth]{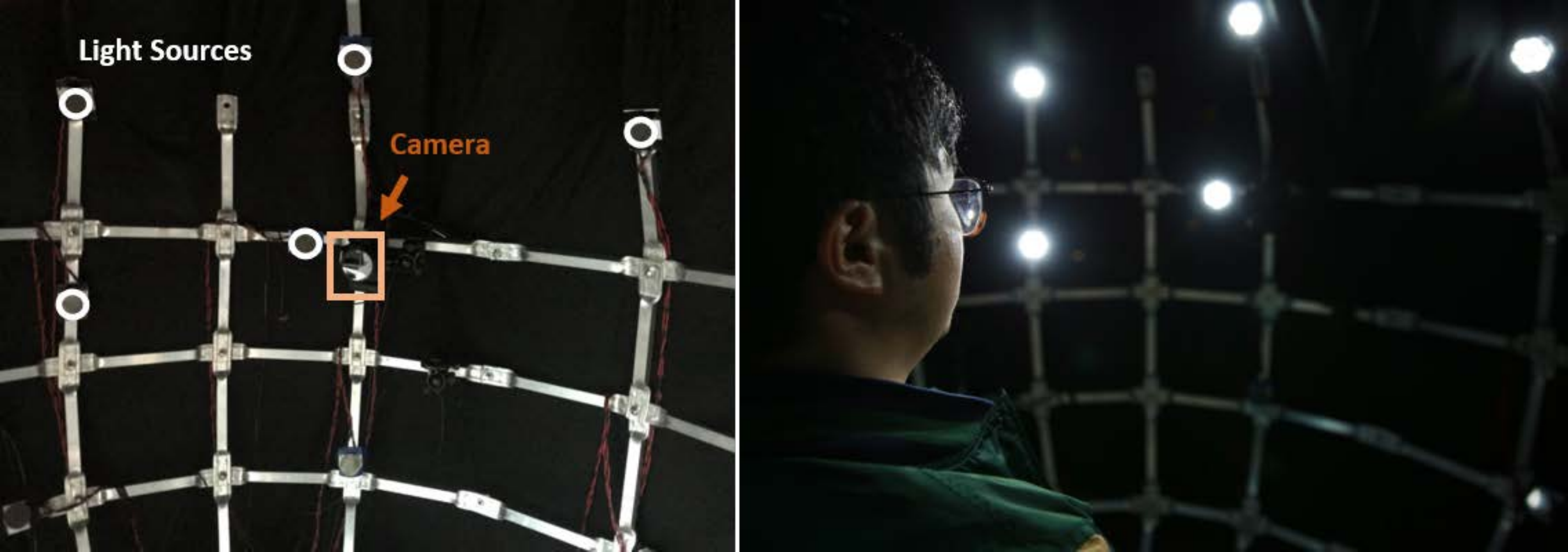}
\caption{We have constructed an acquisition system composed of 5 point light sources and a single DSLR camera.}\label{fig:SystemSetup}
\end{figure}

For real data, we have constructed a sparse photometric capture system composed of $5$ LED near point light sources and an entry-level DSLR camera (Canon 760D) as illustrated in Fig. \ref{fig:SystemSetup}. The distance between the light sources and photographed face is about $1$ meter. To eliminate specular reflectance, both light sources and camera are mounted with polarizers, where the polarizers on light sources are orthogonal to the ones on the camera. Each acquisition captures $5$ images ($1$ light source per image), each at a resolution of 6000x4000. The process takes less than $2$ seconds.

\begin{figure*}[h]
\centering
\includegraphics[width=1\textwidth]{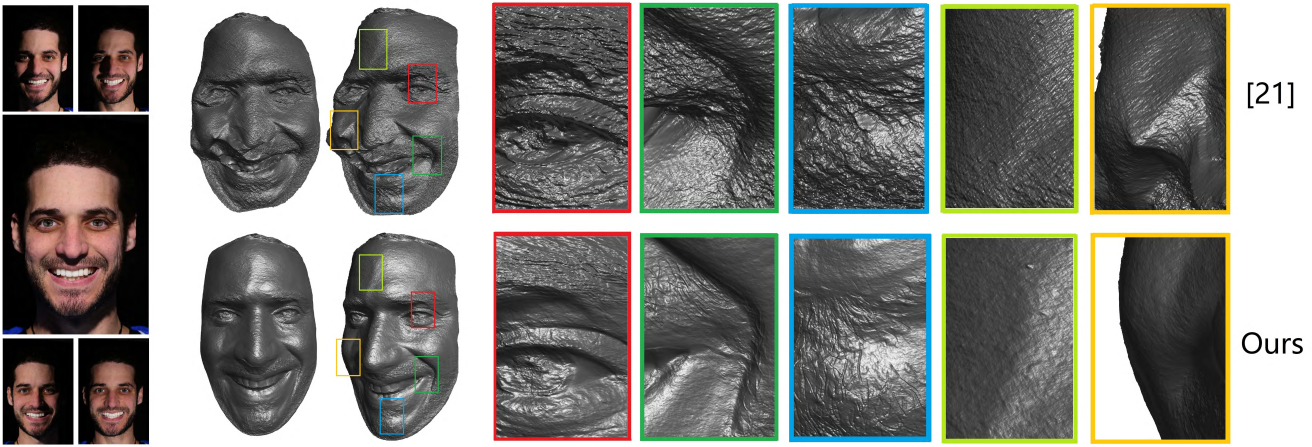}
\caption{Reconstruction results of \cite{park2017robust} (top row) vs. ours (bottom row). \cite{park2017robust} causes large deformations and high noise when using a sparse set of images. Our approach is able to faithfully reconstruct face geometry without deformation and at the same time recover fine details.}\label{fig:CompareFace}
\end{figure*}

\begin{figure*}[t]
\centering
\includegraphics[width=1.0\textwidth]{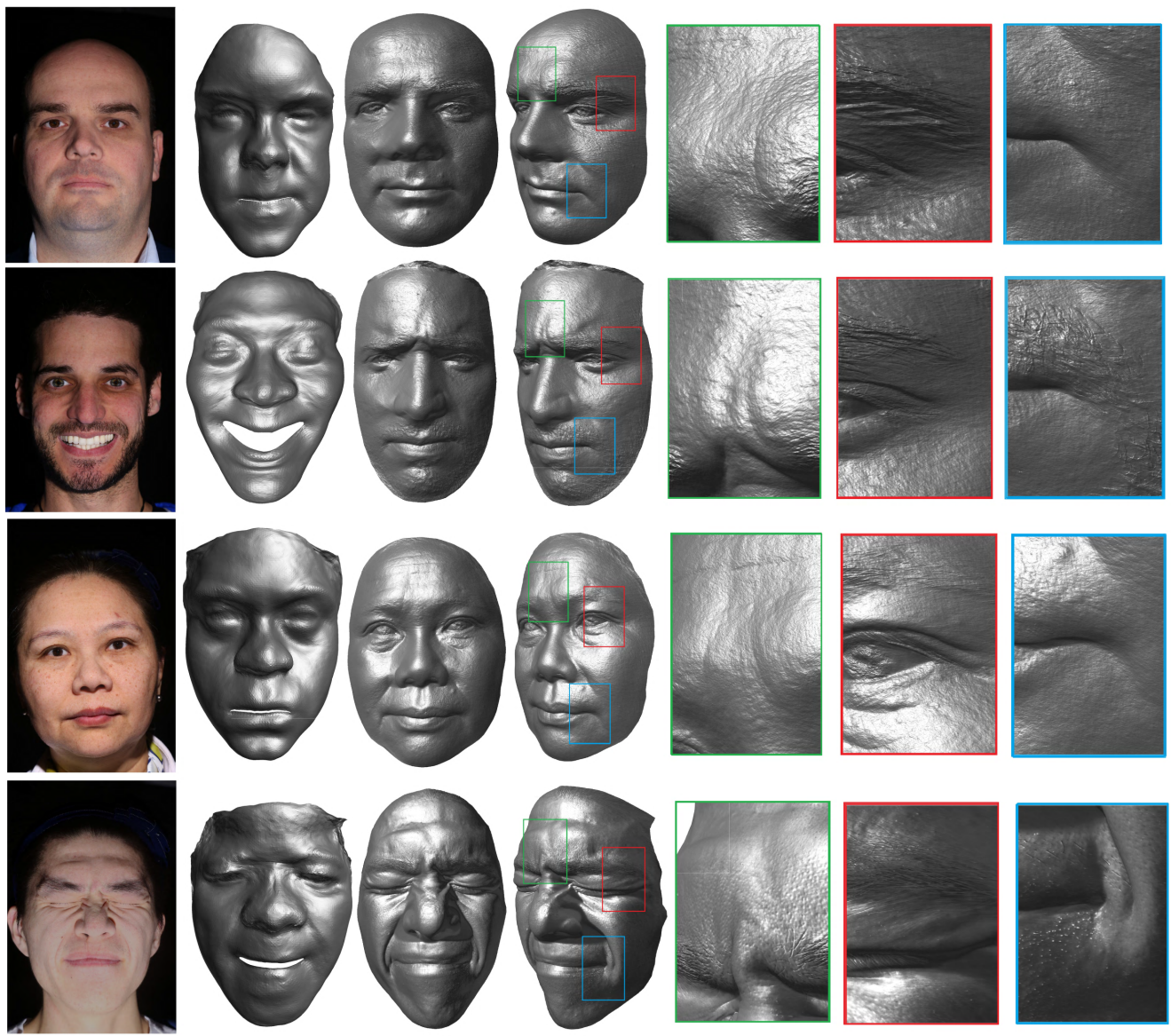}
\caption{Our reconstruction results across gender, race and age. From left to right, we show one of the 5 input images, the proxy face model, and our final reconstruction results. Closeup views of the eyes and mouth regions illustrate fine geometric details recovered by our technique. Additional results can be found in the supplementary materials.}\label{fig:FinalFaceResults}
\end{figure*}

We acquire faces of people with different gender, race and age. Fig. \ref{fig:FinalFaceResults} shows our reconstruction of five faces: the first column shows the proxy models using\cite{huber2016multiresolution}. The model is reasonable but lacks geometric details. Our reconstruction reduces geometric deformations while revealing compelling high-frequency geometric details. We compare our technique with \cite{park2017robust} in Fig. \ref{fig:CompareFace}. Note that neither methods requires using additional instruments for calibration or 3D scanning. The result from \cite{park2017robust} exhibits noisy normals and contains bumpy artifacts over the entire face. In addition, significant geometry deformation emerges on the right cheek. That is mainly because the image contains large areas of shadows that generate significant amount of outliers. The outliers are detrimental to the reconstruction especially only with 5 input images. In contrast, our reconstruction exhibits very high quality and low noise, largely attributed to our optimization techniques together with shadow and hairy region detection schemes, as shown in Fig. \ref{fig:CompareFace}.

\begin{figure}[h]
\centering
\includegraphics[width=0.50\textwidth]{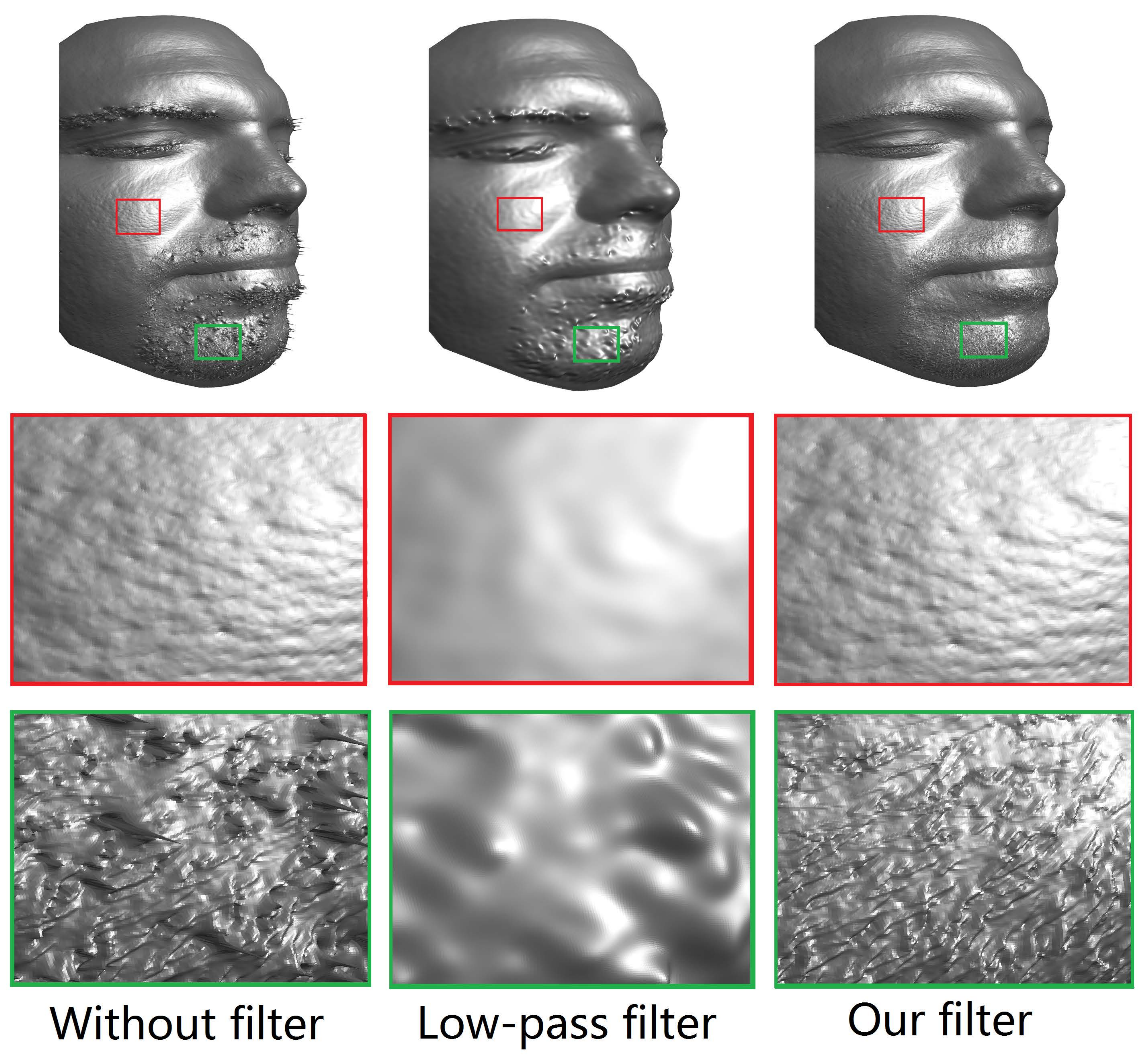}
\caption{Comparisons of different denoising filters in hairy regions. Notice that spiked artifacts in beards are removed by both filters. However, low-pass filter smooth out the high-frequency geometry of hair while our filter preserves such details.}\label{fig:HairDenoise}
\end{figure}

In Fig. \ref{fig:HairDenoise}, we demonstrate the importance and effectiveness of our denoising filters on hairy regions. Without denoising, we observe a large amount of spiking artifacts at the beard and eyebrow regions. Direct low-pass filtering reduces the noise but at the same time over-smooth the geometry. Notice that the beards become roughened after low-pass filtering. Our bidirectional extremum filter, instead, simultaneously removes noise while preserving geometric details. We use the facial region segmentation results in Section \ref{section:Proxy} and only apply our denoising filter on the hairy regions.

\section{Conclusions and Future Work}
We have presented a novel sparse photometric stereo technique for reconstructing very high quality 3D faces with fine details. At the core of our approach is to use base geometric model obtained from morphable 3D faces as geometry proxy for robustly and accurately calibrating the light sources. We have shown our joint optimization strategy is capable of calibration under non-uniform albedo. To fit the base geometry onto the acquired images, we have further presented an iterative reconstruction technique. Finally, we have exploited semantic segmentation techniques for separating hairy vs. bare skin regions where we use bidirectional extremum filters for handling the hairy regions.

Although our paper exploits the 3D morphable face models, we can also potentially use the recent learning-based approaches \cite{Tran2017Regressing,hassner2016pooling} that can produce plausible 3D face models from a single image. In our experiments, we found that the initial result from \cite{Tran2017Regressing}, although visually pleasing, still deviates from the ground truth too much for reliably lighting estimation (see supplementary materials). Our immediate next step therefore is to see how to integrate the shading information into their network framework to produce similar quality results.

There is also an emerging trend of combining semantic labeling with stereo or volumetric reconstruction \cite{hane2017dense,cherabier2016multi}. In our work, we have only used a small set of labels. In the future, we plan to explore more sophisticated semantic labeling technique that can reliably separate a face into finer regions, e.g., eye region, cheek, mouth, teeth, forehead, etc, where we can handle each individual region based on their characteristics. A more interesting problem is how to simultaneously recover multiple faces (of different people) under the photometric stereo setting. For example, if each face exhibits a different pose, a single shot under the directional lighting will produce appearance variations across these faces that are amenable for PS reconstruction.

{\small
\bibliographystyle{ieee}
\bibliography{egbib}
}

\end{document}